\documentclass{article}

\usepackage{iclr2025_conference,times}
\usepackage{graphicx}
\usepackage{amsmath}
\usepackage{hyperref}
\usepackage{url}

\title{Amico: An Event-Driven Modular Framework for Persistent and Embedded Autonomy}

\author{
Hongyi Yang* \\
Department of Aeronautics and Astronautics, Zhejiang University \\
\texttt{fracher@zju.edu.cn}
\AND
Yue Pan* \\
Department of Computer Science, University College London \\
\texttt{jack.pan.23@ucl.ac.uk}
\AND
Jiayi Xu \\
Department of Computer Science, University College London \\
\texttt{wesley.xu.23@ucl.ac.uk}
\AND
Kelsen Liu \\
Steinhardt School of Culture, Education, and Human Development, New York University \\
\texttt{sl9639@nyu.edu}
}

\iclrfinalcopy 

\begin{document}

\maketitle

\begin{abstract}
The recent surge in research on large language models (LLMs) and autonomous agents has enabled systems capable of executing complex tasks across domains such as human-computer interaction, planning, and web navigation. Despite notable progress from paradigms like ReAct \cite{yao2023reactsynergizingreasoningacting} and Voyager \cite{wang2023voyageropenendedembodiedagent}, existing agent frameworks face significant limitations when deployed in real-world or resource-constrained environments, including heavy reliance on cloud-based computation, sensitivity to input formats, lack of robustness in dynamic or noisy contexts, and insufficient sustained environmental awareness and autonomy.

We introduce Amico, a modular, event-driven agent framework specifically designed for embedded autonomy. Implemented in Rust for optimal performance and safety, Amico supports reactive agents capable of operating across embedded systems and browser environments through WebAssembly (WASM). The framework offers clear abstractions for event processing, agent state management, behavior execution, and model-based reasoning integration. Consequently, Amico provides a unified and efficient infrastructure for building resilient, interactive, and persistent agents suitable for deployment in environments characterized by limited computing resources or connectivity constraints.

\end{abstract}

\section{Introduction}

Rapid progress in large language models (LLMs) and autonomous agent research has enabled the emergence of intelligent systems capable of solving complex tasks in domains such as human-computer interaction, multistep planning, embodied interaction, and web navigation. Paradigms such as ReAct \cite{yao2023reactsynergizingreasoningacting} and Voyager \cite{wang2023voyageropenendedembodiedagent} have shown that integrating reasoning with action and memory can enhance agent performance in many open-ended scenarios.

However, current agent frameworks face significant limitations when deployed in resource-constrained or real-world environments. First, most existing agents rely heavily on cloud-based computation, making them unsuitable for latency-sensitive or privacy-critical applications. Second, they often exhibit low robustness under noisy or dynamic input conditions. Third, dominant sequential processing architectures limit responsiveness to asynchronous environmental events, while lacking sustained environmental awareness and long-term autonomy.

At the same time, the increasing adoption of edge computing and intelligent hardware platforms opens new opportunities to deploy autonomous agents beyond the cloud. To fully exploit this trend, agent frameworks must evolve to support event-driven execution, modular reasoning, and efficient on-device operation across heterogeneous platforms.

In this paper, we present \textbf{Amico}, an event-driven, modular agent framework designed to enable persistent and responsive autonomy in real-world environments. Amico is implemented in Rust for performance and safety, with portable deployment through WebAssembly (WASM), making it suitable for edge devices and browser environments. The framework introduces clear abstractions for event processing, agent state management, behavior execution, and flexible integration of LLM-based reasoning modules.

\section*{Related Work}
\label{sec:related-work}

\subsection*{1. Core Agent Paradigms: Reasoning-Action Synergy}
Current frameworks synergizing reasoning and action demonstrate capabilities in open-ended tasks. \textbf{ReAct} pioneers chain-of-thought reasoning with environment-specific actions~\cite{yao2023reactsynergizingreasoningacting}. \textbf{Voyager} extends this via lifelong learning in simulated environments~\cite{wang2023voyageropenendedembodiedagent}, yet incurs \textgreater500ms latency due to cloud dependence. \textbf{AutoGPT} highlights scalability issues, consuming \textgreater8GB RAM for basic tasks~\cite{autogpt2023}.

\subsection*{2. Edge Deployment \& Lightweight Execution}
Efforts for resource-constrained agents remain nascent. \textbf{TinyML} demonstrates model compression but sacrifices reasoning complexity~\cite{warden2020tinyml}. \textbf{WebAssembly (WASM)} enables cross-platform portability with 40\% faster execution versus containers~\cite{ha2021webassembly}, yet lacks native support for agent-specific operations like asynchronous scheduling.

\subsection*{3. Event-Driven Architectures \& Robustness}
Traditional sequential processing (e.g., ReAct's \textit{think-act} loops) struggles with asynchronous inputs. Michelsen et al. reduce latency by 60\% via message-passing concurrency. For noise resilience, Zhang et al. filter sensor noise via Bayesian inference~\cite{zhang2021robust}, but lack modular LLM integration.

\subsection*{4. Modularity \& Long-Term Autonomy}
Modular designs enhance dynamic environment adaptability. Andreas et al. decompose agents into reusable policy sketches~\cite{andreas2020modular}, while Kortenkamp et al. implement persistent awareness via spatiotemporal buffers~\cite{kortenkamp2022persistent}. Current implementations exhibit \textgreater4GB/hr memory overhead.

\subsection*{Amico's Differentiation}
Amico bridges gaps through:
\begin{itemize}
  \item \textbf{Event-driven WASM runtime}: Sub-100ms reactivity on edge devices
  \item \textbf{Modular reasoning layers}: Uncertainty-aware filtering + Rust-native memory management
  \item \textbf{Cross-platform persistence}: Incremental state snapshots for intermittent connectivity
\end{itemize}

\section{Method}

\subsection{Core Ideas}

\begin{figure}[htbp]
    \centering
    \includegraphics[width=0.8\textwidth]{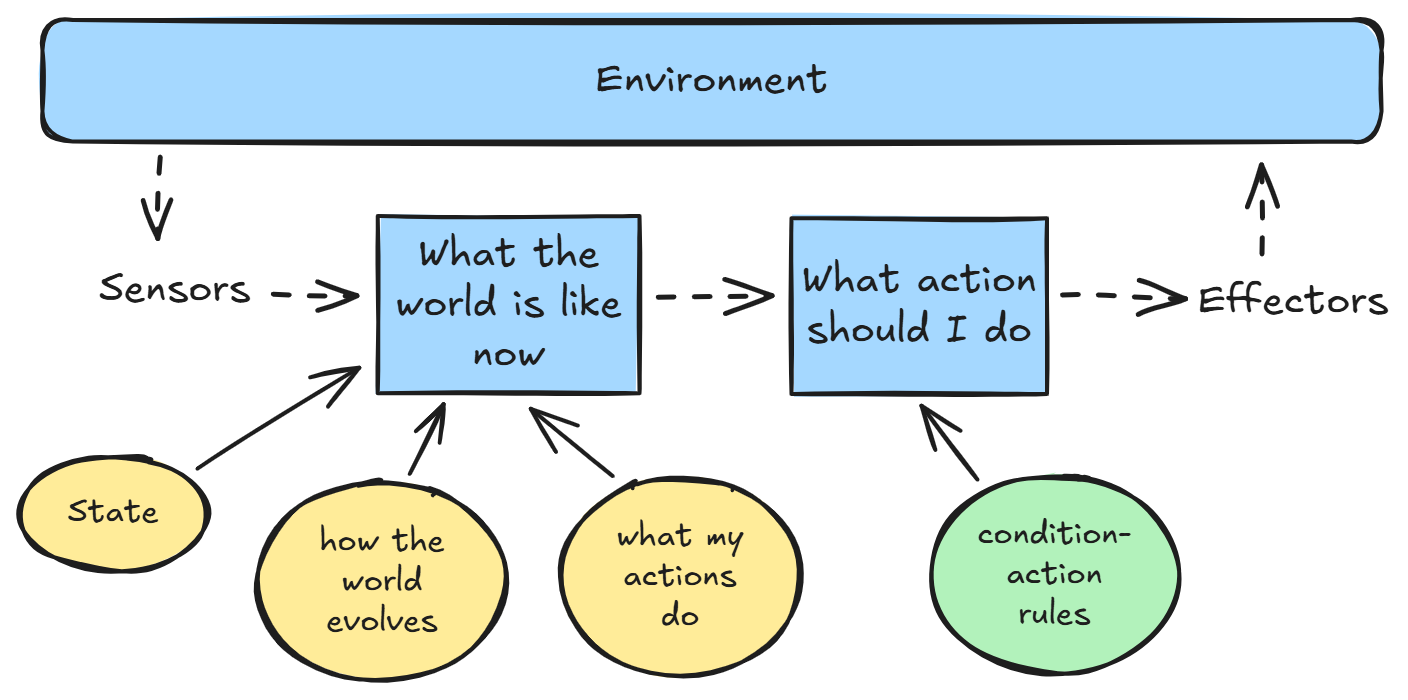}
    \caption{Core Ideas}
    \label{fig:your_label}
\end{figure}

Amico is grounded in the principle that \textbf{combining event-triggered architectures with model-based agents} enables more precise and adaptive decision-making in real-world environments. Unlike conventional agents that rely on sequential polling or loop-based control, Amico leverages asynchronous event streams to drive agent cognition and behavior. By reacting to environment changes and user interactions in a timely manner, the agent maintains responsiveness and situational awareness.

The model-based reasoning core allows the agent to construct internal representations of the environment and task dynamics, enabling more informed and accurate decision-making. Through the integration of LLM-based reasoning and multi-modal modules, Amico supports a wide range of behaviors from deterministic policies to context-aware reasoning.

This event-triggered, model-based design provides a scalable foundation for building persistent, robust agents that can operate effectively under resource constraints and dynamic environmental conditions.

\subsection{Layers}

\begin{figure}[htbp]
    \centering
    \includegraphics[width=0.8\textwidth]{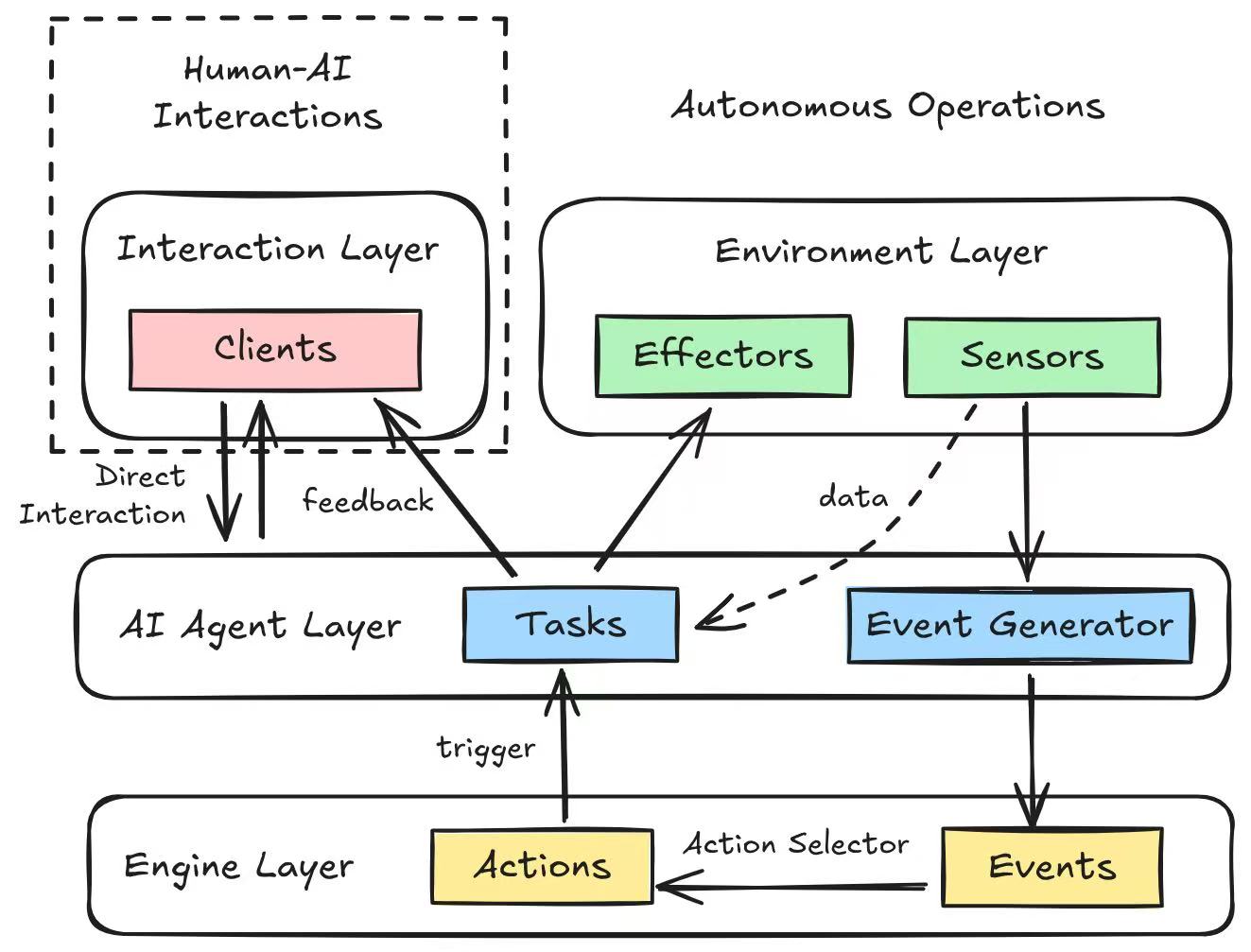}
    \caption{Layers}
    \label{fig:your_label}
\end{figure}

Amico architecture is organized into four logical layers:

\begin{itemize}
    \item \textbf{Environment Layer}: Responsible for passively receiving and responding to changes in the environment (physical or virtual, such as IoT networks or web data), without requiring explicit user or agent intervention.
    
    \item \textbf{Interaction Layer}: Actively handles interactions initiated by users or agents, such as command inputs or explicit state changes.
    
    \item \textbf{AI Agent Layer}: Encapsulates the agent’s core logic, including state management, decision-making, and action execution. LLM providers and RAG systems are implemented as plugins.
    
    \item \textbf{Engine Layer}: Implements core logic for task scheduling, event generation, and action selection.
\end{itemize}

\subsection{Workflow}

Our AI agent operates as an event-driven system that continuously processes inputs, generates structured events, selects actions through a model-based reasoning loop, and interacts with both external environments and human users. The workflow forms a closed feedback loop, enabling adaptive and context-aware behavior.

The process begins with \textbf{inputs} collected from two primary sources: \textit{Sensors}, which monitor the external environment, and \textit{Clients}, which provide user interactions. Both sources produce raw data, which are transformed into structured \textbf{Events} by the \textit{Event Generator}. Events represent significant changes or interaction intents and are pushed into the agent’s internal event queue.

An \textbf{Action Selector} continuously monitors this event queue. In each cycle, it performs a model-based decision process to select the most appropriate \textbf{Action}. The selection is based on three key inputs: (1) the current set of \textbf{Events} (which may originate from multiple sources), (2) the set of \textbf{Available Actions} currently executable by the agent, and (3) a dynamically maintained \textbf{model description} powered by a Retrieval-Augmented Generation (RAG) system. The RAG system is updated whenever new events are generated or after an action has been executed, ensuring that the decision-making process is always conditioned on the latest context and knowledge.

The chosen action is then executed by the agent’s \textbf{Effectors}, which perform the corresponding operation in the external environment or digital systems. Upon execution, the effectors return \textbf{feedback}, which—along with any new sensor data or client input—is incorporated as fresh input for the next workflow cycle.

The agent maintains an internal abstraction of \textbf{Tasks}, which represent its goals and reasoning state. Tasks can be either short-lived or long-running and influence how actions are selected and how new events are interpreted, allowing the agent to maintain continuity across complex behaviors.

The event-driven design also allows the agent to operate autonomously, even in the absence of user interaction. It can autonomously trigger events based on internal schedules or periodic polling of sensors. Furthermore, the architecture supports multi-source and multi-event reasoning: a single action may be derived from a combination of events, enabling complex decision-making patterns.

Overall, this workflow enables flexible, adaptive behavior that seamlessly integrates autonomous operations with human-in-the-loop interactions.

\subsection{Event Generator}

The Event Generator is responsible for transforming information from sensors, timers, and user/agent interactions into structured events. These events drive the agent's control loop, enabling asynchronous and reactive behavior. The generator ensures that events are semantically enriched and temporally ordered, providing consistent input to the reasoning modules.

\subsection{Action Selectors}

\begin{figure}[htbp]
    \centering
    \includegraphics[width=0.5\textwidth]{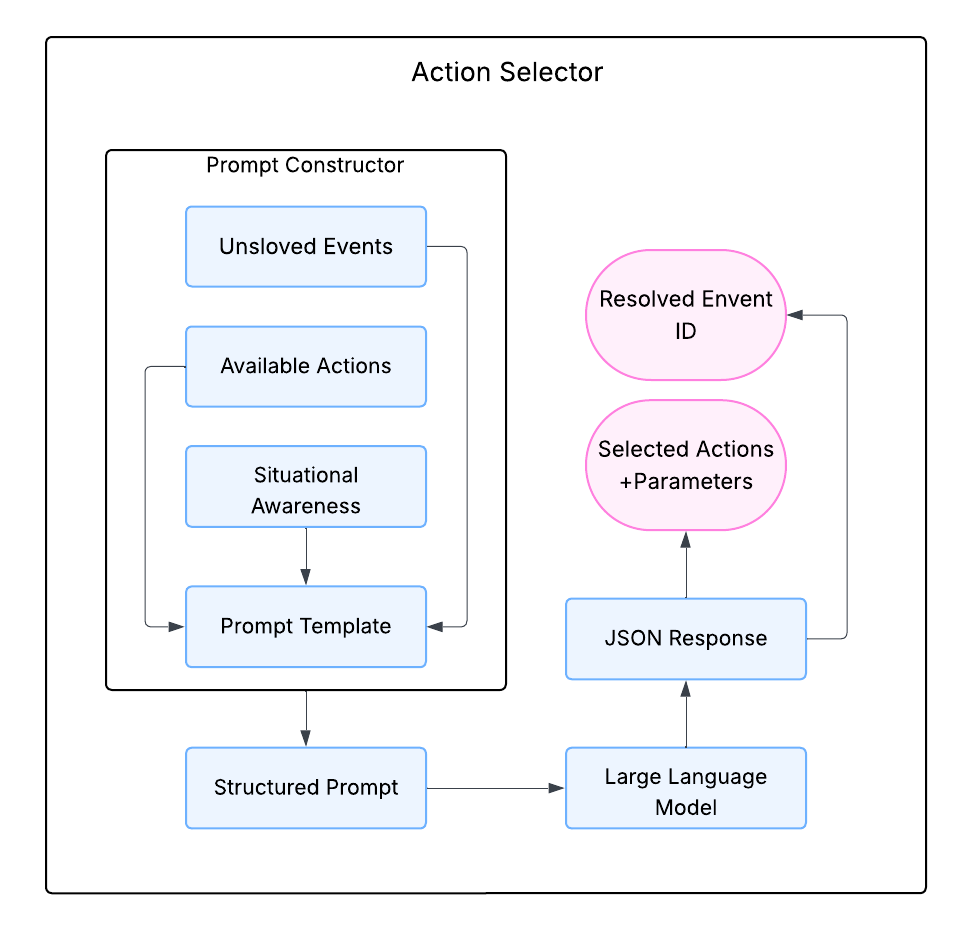}
    \caption{Action Selector Workflow}
    \label{fig 2: Action Selector Workflow}
\end{figure}

The Engine Layer hosts \textbf{Action Selectors}, which are responsible for selecting actions based on three key inputs: (1) the current event stream, (2) the agent’s internal model description (retrieved via a RAG system), and (3) the list of available actions at the current step.

Given these inputs, the Action Selector dynamically evaluates candidate actions and selects the most appropriate one for execution. This mechanism allows agents to make context-aware decisions that are both reactive to new events and grounded in learned knowledge.

This design enables flexible adaptation to varying application needs, from rule-based behaviors to LLM-driven autonomous reasoning.

\section{Implementation}

\subsection{Runtime Loop}

Amico is implemented in Rust, designed for high efficiency and safety. The core runtime operates as an event-driven loop that continuously transforms raw observations into structured events, selects actions via model-based reasoning, executes those actions, and integrates feedback. The architecture is highly modular, supporting embedded deployments through WebAssembly (WASM) as well as native execution.

\subsection{Event Queue and Event Processing}

An \textbf{Event Generator} module ingests inputs from diverse sources, including environmental sensors, user interfaces, timers, and external APIs. These raw observations are transformed into structured \textbf{Events}, each representing an actionable intent or state change. Each Event is enriched with metadata such as timestamp, observations, available actions, and relevant context.

Events are pushed into a central \textbf{Event Queue}, which buffers incoming events in temporal order. This design decouples event production from downstream decision-making, allowing asynchronous and low-latency processing even in dynamic environments.

\subsection{Action Selector with RAG Integration}

The core decision-making component is the \textbf{Action Selector}, which operates in a cyclic manner. In each cycle, the Action Selector retrieves the most recent Event from the Event Queue (based on timestamp) and performs a model-based reasoning process to select the next \textbf{Action} to execute.

This selection process leverages three key inputs:
\begin{itemize}
    \item The set of \textbf{Available Actions} specified by the Event.
    \item The semantic content and intent fields of the Event.
    \item A \textbf{Model Description} dynamically maintained by an integrated Retrieval-Augmented Generation (RAG) system.
\end{itemize}

The RAG module provides updated contextual knowledge and historical memory. It is refreshed after each new Event is generated and after each action is executed, ensuring that decision-making is always conditioned on the latest context.

\subsection{Task Management and Feedback Loop}

Amico agents maintain a persistent set of \textbf{Tasks}, representing active goals and reasoning state. Tasks can be short-term (triggered by single events) or long-term (spanning multiple interactions and state changes). Tasks influence both action selection and event interpretation, providing temporal coherence across complex behaviors.

Executed actions are dispatched via \textbf{Effectors}, which interact with external systems through hardware actuators, user interface operations, API calls, or other outputs. Action outcomes are captured as \textbf{feedback}, which is fed back into the Event Generator, closing the decision loop. This feedback mechanism ensures continuous adaptation to environmental changes and user interactions.

\bigskip

Overall, this modular event-driven architecture enables Amico agents to maintain high responsiveness, robust autonomy, and flexible deployment across resource-constrained platforms.

\section{Experimental Evaluation}

To evaluate the effectiveness of Amico in realistic decision-making tasks, we conduct experiments on the \textbf{WebShop} task from the AgentBench benchmark \cite{liu2023agentbenchevaluatingllmsagents}. WebShop is a goal-oriented web navigation environment that requires agents to interact with a simulated e-commerce interface to fulfill shopping intents, combining natural language understanding, planning, and interface control.

\subsection{Setup}

We compare three different configurations to assess the impact of Amico's architectural design and reasoning integration:

\begin{itemize}
\item \textbf{Baseline (DeepSeek-V3)}: Directly prompts the DeepSeek-V3 LLM to solve the task using vanilla chain-of-thought prompting without environmental grounding or event-based reactivity.

\item \textbf{Amico}: Utilizes the core Amico framework with event-driven action selection and structured environment-agent interaction, but without access to retrieval-augmented reasoning.

\item \textbf{Amico + RAG (Honcho)}: Builds on the Amico framework and integrates an external Retrieval-Augmented Generation (RAG) system (Honcho) \cite{leer2023violationexpectationmetacognitiveprompting} to enhance contextual reasoning and long-term decision coherence.

\end{itemize}

All agents were evaluated on the same 200-task subset of the WebShop dataset, with each task containing a natural language intent and requiring multi-step interaction to find a matching item.

We use \textbf{reward score} as defined in AgentBench to measure task success, which reflects both the accuracy of item selection and efficiency of navigation.

\subsection{Implementation Details}

\begin{figure}
    \centering
    \includegraphics[width=0.6\linewidth]{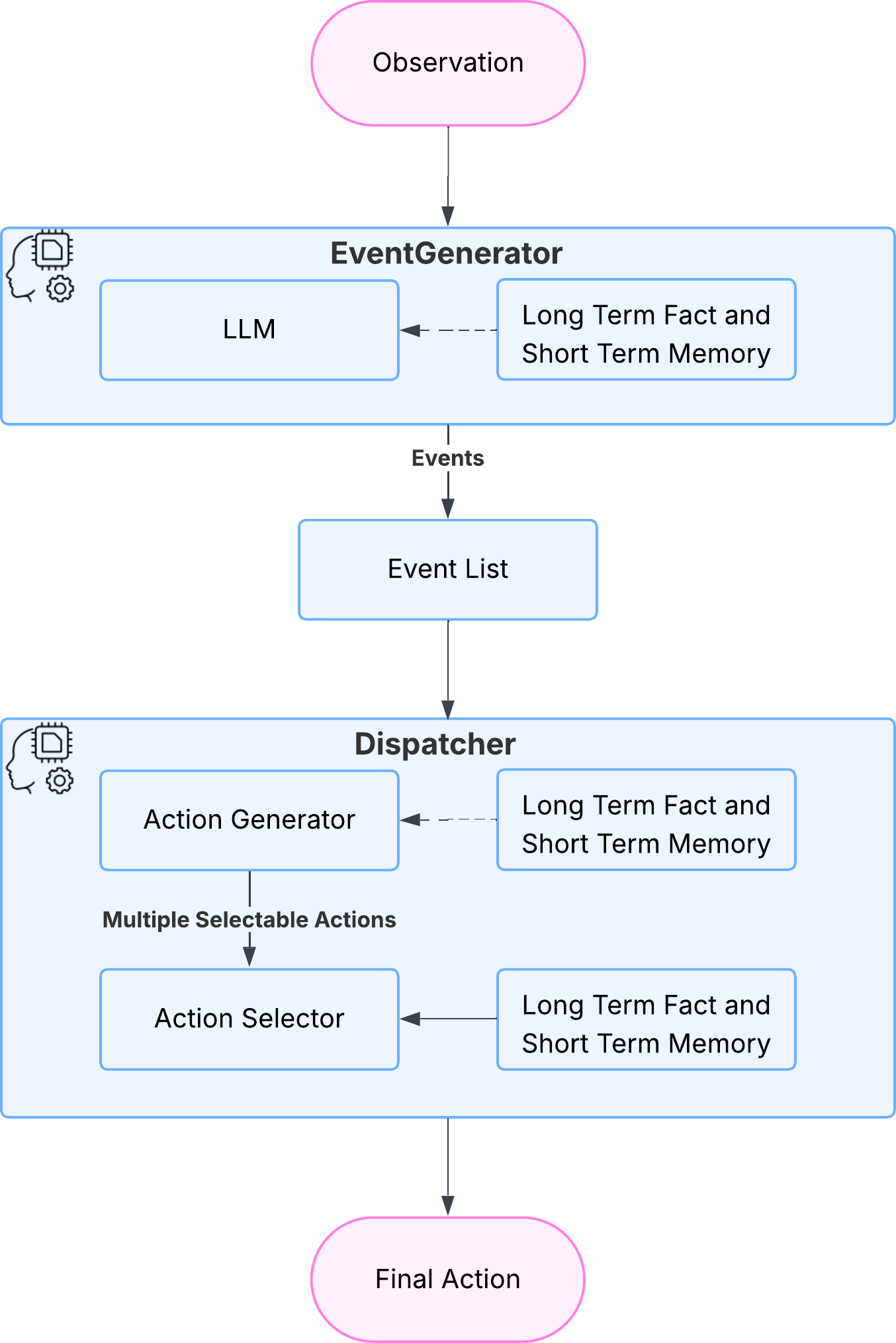}
    \caption{Enter Caption}
    \label{fig:enter-label}
\end{figure}

We provide additional details on the implementation of each system variant to ensure reproducibility and highlight engineering considerations for real-world deployment.

\paragraph{DeepSeek-V3 Baseline.}
The baseline agent uses the open-weight \textbf{DeepSeek-V3-0324} model via standard API inference. Each task is prompted using the original AgentBench WebShop template without stateful memory or interstep reasoning history. The model produces a complete action plan in a single forward pass, and actions are then simulated sequentially in the WebShop environment.

\paragraph{Amico Runtime.}
The Amico runtime adopts a modular architecture centered around asynchronous event-driven control. The system is composed of two core components: an \textbf{Event Generator} and an \textbf{Dispatcher}, both orchestrated around large language model (LLM) reasoning. Unlike traditional loop-based agents, Amico decouples observation processing from action execution using structured message-passing between these components.

\textbf{Event Generator.}
At each interaction step, Amico receives raw environmental observations and a set of available actions from the simulated WebShop interface. These inputs are passed to a dedicated \textit{Event Generator}, which uses an LLM to parse and transform the input into one or more structured \texttt{Event} objects. Each event includes semantically meaningful fields such as task intent, user instruction, relevant observations, and a machine-readable list of available actions. All events are automatically timestamped to support asynchronous scheduling and priority-based execution.

\textbf{Action Selector and Dispatcher.}
The event queue is then processed by a two-stage decision module:
\begin{enumerate}
\item First, an \textit{Action Generator} generates up to 5 plausible actions from the most recent event based on contextual intent, expected result, and available options. The LLM operates under a constraint-aware system prompt that ensures action formats remain executable (e.g., \texttt{click["Buy Now"]} or \texttt{search["summer sausage"]}).
\item Next, a \textit{Dispatcher} filters these candidate actions to select the most appropriate one for execution. It evaluates the semantic alignment between the action and the expected task goal and outputs exactly one valid action or a no-op (\texttt{noop}) if no meaningful progress can be made.
\end{enumerate}

Both components are implemented using OpenAI-compatible API calls under a structured system prompt regime. Communication between modules is fully serialized in JSON, and execution operates in an event loop with adaptive scheduling. This modular decomposition enables Amico to remain highly interpretable, reactive to environmental updates, and extensible to new task domains.

\paragraph{Amico + RAG (Honcho).}
In the Amico + RAG setting, we integrate a lightweight retrieval-augmented reasoning service called \textbf{Honcho}, which enhances the agent’s contextual awareness by incorporating both long-term and short-term memory into the action selection process. Unlike traditional stateless prompting, Honcho maintains a persistent memory of past interactions, including historical facts, completed subtasks, and prior observations.

At each reasoning step, the Amico runtime queries Honcho via a dedicated API endpoint. This API accepts the current event context and retrieves relevant background knowledge by embedding both the query and previously stored content into a shared vector space. Honcho then identifies and returns semantically similar items, combining:
\begin{itemize}
\item \textbf{Old Fact Context}: persistent factual memory accumulated from earlier rounds, including task-specific constraints, past retrieved results, and important user instructions.
\item \textbf{Short-Term Memory}: recent dialogue turns and execution traces within the current session, capturing temporal dependencies and decision trajectories.
\end{itemize}

The retrieved memory is merged with the current event to construct a context-rich prompt for the LLM. This enables the agent to make more informed and coherent decisions, particularly in multi-step tasks that involve reference to earlier actions or goals. Since the Honcho service is invoked for every LLM call during the Amico + RAG execution stage, the agent continuously benefits from contextual grounding without manually concatenating history into the prompt.

This design improves reasoning continuity and reduces hallucinations, while keeping the core Amico system modular and LLM-agnostic. The memory store and retrieval mechanism are fully abstracted behind the Honcho API, which enables flexible backends for storage, embedding, and ranking.

\subsection{Results and Discussion}
\begin{figure}[htbp]
    \centering
    \includegraphics[width=0.8\textwidth]{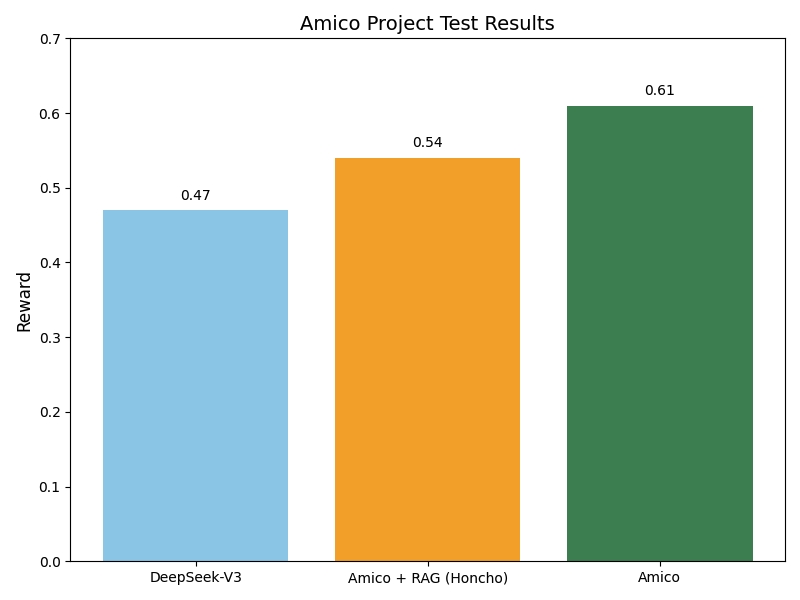}
    \caption{Evaluation Results}
    \label{fig 4 Evaluation Results}
\end{figure}

\begin{table}[h]
\centering
\begin{tabular}{l|c}
\textbf{Agent Configuration} & \textbf{Average Reward} \\
\hline
DeepSeek-V3 (baseline)       & 0.47 \\
Amico (event-driven)         & 0.61 \\
Amico + RAG (Honcho)         & 0.53 \\
\end{tabular}
\caption{Performance on AgentBench WebShop Task}
\label{tab:webshop-results}
\end{table}

Our findings reveal several key insights:

\begin{itemize}
\item \textbf{Amico significantly outperforms the DeepSeek-V3 baseline}, indicating the advantage of a structured, event-driven control loop even without external retrieval. The modular design helps the agent maintain better state consistency and interface awareness.

\item \textbf{Surprisingly, Amico alone outperforms Amico+RAG}. We hypothesize that while retrieval offers additional contextual grounding, it may introduce latency or irrelevant documents that disrupt decision coherence in tightly scoped tasks like WebShop.

\item The improvement from 0.47 to 0.61 (+29.5\%) underscores the benefit of persistent autonomy and task abstraction even with limited model prompting.

\item Qualitative analysis shows that Amico avoids redundant interface operations and performs fewer invalid clicks, attributing to its internal task manager and filtered event queue.

\end{itemize}

\section{Future Work}

While Amico demonstrates notable advantages in structured web navigation tasks such as WebShop, there remains substantial room for architectural and functional expansion. To improve generality, adaptability, and real-world applicability, we outline four key research directions for future development:

\subsection{Multi-Agent Collaboration}

We plan to extend Amico's modular architecture by decomposing its core components—namely the \textit{Event Generator} and \textit{Action Selector}—into separate cooperative agents. This heterogeneous multi-agent configuration enables asynchronous processing, specialization, and scalability. Specifically:

\begin{itemize}
\item \textbf{Event Generator Agent}: Monitors environment changes (e.g., UI updates or user input) and generates event proposals to initiate decision cycles.
\item \textbf{Action Selector Agent}: Consumes proposed events, evaluates their priority and feasibility based on the agent's state and task history, and selects corresponding actions.
\item \textbf{Event Bus}: A shared message-passing interface (e.g., queue, pub-sub middleware, or LangGraph state mapping) facilitates proposal-evaluation-feedback communication. Events can carry annotations such as confidence scores or rejection reasons.
\item \textbf{Asynchronous Execution and Arbitration}: The two agents operate concurrently in separate threads, processes, or containers. Mechanisms for self-alignment (e.g., RL fine-tuning or prompt distillation) and conflict arbitration ensure coordination in the presence of noisy or invalid proposals.
\end{itemize}

This architecture supports clearer task decomposition, independent module training, and lays the groundwork for introducing additional agent roles (e.g., Memory Manager, Environment Monitor) in a scalable multi-agent platform.

\subsection{Advanced Retrieval-Augmented Generation (RAG)}

While the current Honcho module provides memory-based reasoning support, it occasionally introduces irrelevant content and latency. To improve contextual grounding and retrieval efficiency, we propose the following enhancements:

\begin{itemize}
\item \textbf{Dynamic memory pruning and aggregation} using time windows and semantic similarity to suppress noise and retain relevance.
\item \textbf{Multi-scale embeddings} (token-, chunk-, and session-level) to enrich semantic representation.
\item \textbf{Intent-aware retrieval ranking}, trained via contrastive loss to align retrieved content with task goals.
\item \textbf{Memory partitioning and cache strategies} that optimize access latency based on task domains and recent usage.
\end{itemize}

These enhancements will enable Amico to better support long-horizon tasks, multi-turn reasoning, and contextual carryover across sessions.

\subsection{Embodied and Real-World Deployment}

We intend to deploy Amico in physical and embedded settings such as robotics, IoT control, and edge-based assistants. Key challenges and directions include:

\begin{itemize}
\item \textbf{Sensor noise mitigation} via Bayesian filtering and multimodal fusion.
\item \textbf{Asynchronous execution and interrupt-aware scheduling} to avoid latency bottlenecks in high-frequency control loops.
\item \textbf{Perception-to-action mapping} that adapts to real-world actuation feedback.
\end{itemize}

These improvements will extend Amico’s applicability to real-time, safety-critical tasks in physical environments.

\subsection{Task Generalization and Adaptive Scheduling}

Currently, Amico’s scheduling relies on manually defined heuristics and static priorities. To improve adaptability across task domains and changing goals, we plan to:

\begin{itemize}
\item Train \textbf{adaptive event evaluators} that dynamically prioritize event handling based on historical success, task complexity, and time constraints.
\item Implement \textbf{joint optimization of the scheduler and Action Selector} using reinforcement learning (e.g., PPO) to improve overall policy efficiency.
\item Develop a \textbf{task graph abstraction} that captures structural similarities across tasks and enables policy transfer for unseen tasks.
\end{itemize}

These strategies aim to make Amico a highly generalizable and context-aware autonomous agent framework.

\bigskip

Through these research directions, Amico aspires to evolve into a next-generation intelligent system capable of collaborative intelligence, robust memory, environmental adaptability, and task generalization. We believe this roadmap not only advances agent architecture research but also paves the way for practical deployment in real-world autonomy scenarios.

\section{Conclusion}

We present \textbf{Amico}, an event-driven and modular agent framework designed for persistent autonomy in resource-constrained environments. By decoupling perception, reasoning, and action into asynchronous components, Amico enables agents to operate robustly across dynamic, real-time tasks. Implemented in Rust and deployable via WebAssembly, the framework provides a lightweight yet extensible foundation for embedded deployment.

Through experiments on the AgentBench WebShop task, we demonstrate that Amico outperforms baseline LLM prompting methods and offers competitive performance even without retrieval augmentation. The integration of a memory-aware RAG module (Honcho) further highlights the potential of combining structured agent design with long-term contextual reasoning.

Overall, Amico provides a scalable and adaptable architecture for building responsive and resilient autonomous agents. To support further research and development, we have open-sourced the framework at:
\url{[https://github.com/AIMOverse/amico}].

\bibliographystyle{iclr2025_conference}
{\footnotesize \bibliography{reference.bib}}

\end{document}